\documentclass[a4paper,fleqn]{cas-dc}

\usepackage[numbers, sort]{natbib}
\def\tsc#1{\csdef{#1}{\textsc{\lowercase{#1}}\xspace}}
\tsc{WGM}
\tsc{QE}
\tsc{EP}
\tsc{PMS}
\tsc{BEC}
\tsc{DE}
\usepackage{amsmath,graphicx,mathrsfs,booktabs,type1cm,amssymb,threeparttable,multirow, booktabs, changepage}

\usepackage{footnote}
\usepackage{color, xcolor, soul}
\usepackage{verbatim}
\usepackage{bm}
\usepackage{url}
\usepackage{array}
\usepackage{stfloats}
\usepackage{float}
\usepackage{graphicx}
\usepackage{makecell, rotating}
\usepackage{threeparttable}
\usepackage{textcomp}
\usepackage[caption=false, justification=centering]{subfig}
\usepackage{enumitem}

\usepackage{media9}

\begin{document}
\begin{sloppypar}
\let\WriteBookmarks\relax
\def\floatpagepagefraction{1}
\def\textpagefraction{.001}

\shorttitle{Retrieval-Augmented Dialogue Knowledge Aggregation for Expressive Conversational Speech Synthesis}

\shortauthors{Rui Liu et~al.}

\title [mode = title]{Retrieval-Augmented Dialogue Knowledge Aggregation for Expressive Conversational Speech Synthesis}                      
% \tnotemark[1,2]

% \tnotetext[1]{This document is the results of the research
%    project funded by the National Science Foundation.}

% \tnotetext[2]{The second title footnote which is a longer text matter
%    to fill through the whole text width and overflow into
%    another line in the footnotes area of the first page.}

\author[1]{Rui Liu}[type=editor,
                        % auid=000, bioid=1,
                        % prefix=Sir,
                        % role=Researcher,
                        % orcid=0000-0001-7511-2910
                        ]

% \cormark[1]


\ead{liurui_imu@163.com}

% \ead[url]{www.cvr.cc, cvr@sayahna.org}

\credit{Conceptualization, Methodology, Formal analysis,
Data curation, Writing – original draft, Writing – review & editing, Validation, Funding acquisition.}

\affiliation[1]{organization={Inner Mongolia University},
    % addressline={235}, 
    city={Hohhot},
    % citysep={}, % Uncomment if no comma needed between city and postcode
    postcode={010021}, 
    country={China}}

\affiliation[2]{organization={The Chinese University of Hong Kong},
    % addressline={235}, 
    city={Shenzhen},
    % citysep={}, % Uncomment if no comma needed between city and postcode
    postcode={999077}, 
    country={China}}

\author[1]{Zhenqi Jia}[style=chinese]
\credit{Methodology, Formal analysis,
Data curation, Investigation, Resources}
\author[1]{Feilong Bao}[%
   % role=Co-ordinator,
   % suffix=Jr,
   ]
   
\author[2]{Haizhou Li}[%
   % role=Co-ordinator,
   % suffix=Jr,
   ]
% \ead{csggl@imu.edu.cn}
% \ead[URL]{www.sayahna.org}

\credit{Methodology,
Supervision.}

% \affiliation[2]{organization={Sayahna Foundation},
%     % addressline={}, 
%     city={Jagathy},
%     % citysep={}, % Uncomment if no comma needed between city and postcode
%     postcode={695014}, 
%     state={Trivandrum},
%     country={India}}

% \affiliation[3]{organization={STM Document Engineering Pvt Ltd.},
%     addressline={Mepukada}, 
%     city={Malayinkil},
%     % citysep={}, % Uncomment if no comma needed between city and postcode
%     postcode={695571}, 
%     state={Trivandrum},
%     country={India}}

% \cortext[cor1]{Corresponding author}
% \cortext[cor2]{Principal corresponding author}

% % Footnote text
% \fntext[fn1]{This is the first author footnote. but is common to third
% \fntext[fn2]{Another author footnote, this is a very long footnote and
%   it should be a really long footnote. But this footnote is not yet
%   sufficiently long enough to make two lines of footnote text.}

% \nonumnote{xxxx}

\nonumnote{  Corresponding Author: Feilong Bao }

\begin{abstract}
Conversational speech synthesis (CSS) aims to take the current dialogue (CD) history as a reference to synthesize expressive speech that aligns with the conversational style. Unlike CD, stored dialogue (SD) contains preserved dialogue fragments from earlier stages of user-agent interaction, which include style expression knowledge relevant to scenarios similar to those in CD.
 Note that this knowledge plays a significant role in enabling the agent to synthesize expressive conversational speech that generates empathetic feedback. However, prior research has overlooked this aspect. To address this issue, we propose a novel \textbf{R}etrieval-\textbf{A}ugmented \textbf{D}ialogue \textbf{K}nowledge \textbf{A}ggregation scheme for expressive CSS, termed \textbf{RADKA-CSS}, which includes three main components: 1) To effectively retrieve dialogues from SD that are similar to CD in terms of both semantic and style. First, we build a stored dialogue semantic-style database (SDSSD) which includes the text and audio samples.
 %  based on DailyTalk, which includes the text and audio content of each dialogue, as well as dialogue semantic and style vectors. 
 Then, we design a multi-attribute retrieval scheme to match the dialogue semantic and style vectors of the CD with the stored dialogue semantic and style vectors in the SDSSD, retrieving the most similar dialogues.
 % based on combined semantic and style similarity.
 2) To effectively utilize the style knowledge from CD and SD, we propose adopting the multi-granularity graph structure to encode the dialogue and introducing a multi-source style knowledge aggregation mechanism. 
 3) Finally, the aggregated style knowledge are fed into the speech synthesizer to help the agent synthesize expressive speech that aligns with the conversational style. 
 We conducted a comprehensive and in-depth experiment based on the  DailyTalk dataset, which is a benchmarking dataset for the CSS task.
 Both objective and subjective evaluations demonstrate that RADKA-CSS outperforms baseline models in expressiveness rendering. Code and audio samples can be found at: {\url{https://github.com/Coder-jzq/RADKA-CSS}}.

\end{abstract}

% \includegraphics{figs/grabs.pdf}

\begin{highlights}
\item We propose a novel Retrieval-Augmented Dialogue Knowledge Aggregation CSS model, termed RADKA-CSS.
     
\item We introduce a dialogue semantic-style database and design a multi-attribute retrieval module to facilitate style- and semantics-based dialogue retrieval.

\item We propose a multi-granularity heterogeneous graph modeling mechanism that aims to capture the structural and temporal relations of nodes at different granularities, effectively encoding dialogue semantics and style information.

\item The experimental results on the DailyTalk benchmarking dataset validate the effectiveness of RADKA-CSS.

\end{highlights}

% Keywords

\maketitle

\section{Introduction}
\label{sec:introduction}
Conversational speech synthesis (CSS) \cite{guo2021conversational} aims to use current dialogue (CD) history as a reference to express a target utterance with the proper linguistic and affective
prosody in a user-agent conversational context \cite{liu2024emotion}. In the user-agent interaction system, the ability of the agent to synthesize speech that aligns with the current conversational style is vital for user experience \cite{deng2023cmcu}. As human-computer interaction (HCI) becomes increasingly prevalent, CSS has become a crucial component of intelligent interactive systems \cite{zhou2020design, seaborn2021voice, mctear2022conversational} and plays an important role in areas such as virtual assistants, voice agents, etc.

\begin{figure}
    \centering
    \vspace{2mm}\includegraphics[width=1\linewidth]{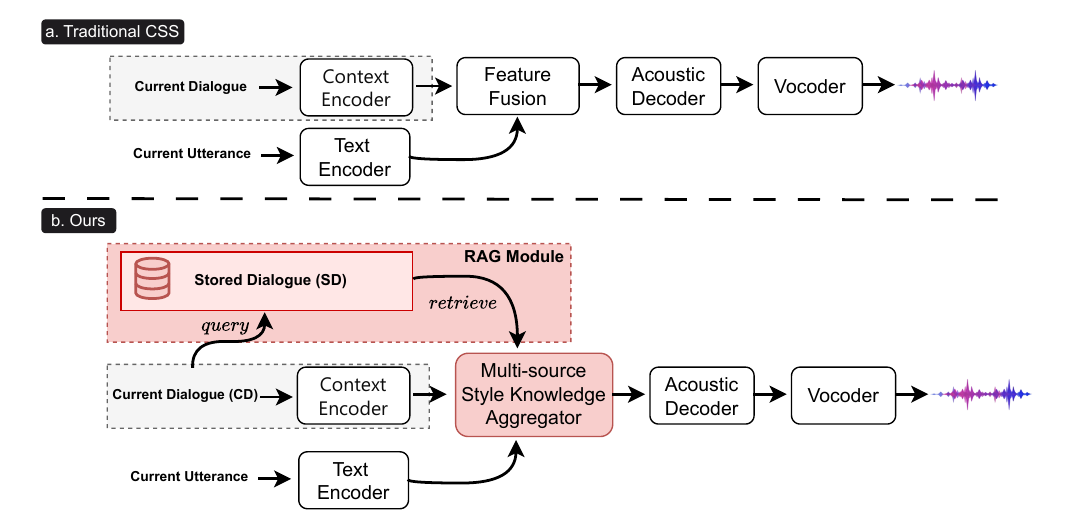}
    % \vspace{-3mm}
    \caption{From traditional CSS to our proposed retrieval-augmented dialogue knowledge aggregation CSS.}
        % \vspace{-3mm}
    \label{ragcss}
\end{figure}

Unlike text-to-speech task \cite{shen2018natural, ren2021fastspeech, liu23u_interspeech, zhu2024unistyle, liu2024controllable, liu2024text, shen2024naturalspeech}, the traditional CSS task mainly focuses on CD context modeling, as shown in Fig. \ref{ragcss}(a), which can be summarized into three groups: \textbf{1) Multi-scale context modeling}, \cite{guo2021conversational} introduces a GRU-based coarse-grained context encoder that extracts semantic information from sentence-level historical dialogues. \cite{hu2024fctalker} further considers simultaneously learning coarse-grained and fine-grained contextual dependencies from text. \cite{li2022inferring} infers speaking styles in dialogues using a multi-scale relational graph convolutional network. \textbf{2) Multi-modal context modeling}, \cite{nishimura22_interspeech} demonstrates that combining acoustic features with textual semantic information improves speech synthesis quality. \cite{xue2023m} integrates coarse-grained and fine-grained modeling of dialogue history from both text and audio modalities. \cite{deng2024concss} models the discriminability of text and audio contextual features through a contrastive learning module. \cite{liu2024emotion} constructs multi-source knowledge from dialogue history into a heterogeneous graph for emotionally expressive speech synthesis. \textbf{3) Spontaneous behaviors modeling}, \cite{cong21b_interspeech} uses a speaker-independent acoustic context encoder and a BERT-based predictor to model spontaneous behavior in conversations. \cite{mitsui22_interspeech} proposes an end-to-end speech synthesis method for spontaneous conversations, using a two-stage training approach to model and predict speaking style based on dialogue history.

However, previous CSS work \cite{ guo2021conversational, o2022combining, xue2023m, deng2023cmcu, deng2024concss, liu2024emotion, mitsui22_interspeech, cong21b_interspeech, mei2024considering, liu2024generative} just focused on modeling CD history. 
 Unlike CD, stored dialogue (SD) contains preserved dialogue fragments from earlier stages of user-agent interaction, which include style expression knowledge relevant to scenarios similar to those in CD \cite{lewis2020retrieval}. Fully leveraging this knowledge helps the agent better understand the conversational style of CD \cite{lewis2020retrieval, wang2024unims, xue24b_interspeech}. For example, \cite{bang2015example, wang2024unims} enhances personalized expression by selecting the most similar cases from SD. Therefore, CSS can be considered a knowledge-intensive task that relies on SD. In user-agent interaction, for the agent to generate natural speech that aligns with the conversational style, it needs to reference numerous similar dialogue scenarios. Additionally, by expanding SD dialogue fragments, the agent can access richer dialogue content, thereby improving its ability to adapt to any dialogue scenario.

Retrieval-augmented generation (RAG) methods provide an excellent example of addressing knowledge-intensive tasks \cite{gao2023retrieval, lewis2020retrieval, guu2020retrieval}. Fundamentally, the RAG framework compensates for the limitations of a single model's knowledge by leveraging significant, real-time, and additional explainable knowledge. This enables the model to provide more reasonable answers, with the supplementary contextual information allowing the single model to overcome its inherent knowledge deficiencies \cite{kang2024retrieval, yuan2024retrieval, yang2024rag, ghosh2024recap, xue24b_interspeech}. In user-agent interaction, this approach is equally applicable. By introducing rich SD knowledge, the agent can access the most recent and relevant style knowledge, generating speech that better aligns with the conversational style and enhancing the overall interaction experience \cite{xue24b_interspeech}. As shown in Fig. \ref{ragcss}(b), our model incorporates an RAG module into the CSS framework. The RAG module retrieves dialogues similar to the current conversational style from SD, aggregating style knowledge from both CD and SD, providing the agent with a broad reference that enables it to adapt more effectively to various conversational scenarios.

To address this issue, we propose a novel \textbf{R}etrieval-\textbf{A}ugmented \textbf{D}ialogue \textbf{K}nowledge \textbf{A}ggregation scheme for expressive CSS, termed \textbf{RADKA-CSS}, which includes three main components: 1) To effectively retrieve dialogues from SD that resemble CD in terms of scenarios and style. In the first stage, we build a stored dialogue semantic-style database (SDSSD) based on DailyTalk, which includes the text and audio content of each set of dialogues, as well as the semantic and style vectors of the dialogues. In the second stage, we design a multi-attribute retrieval method. First, to avoid incomplete dialogue style information due to the lack of an $a_N$ style vector ($a_N$ represents the speech to be synthesized, which is unavailable during inference, defined in Section 3), we design an $a_N$ vector predictor to predict the style vector for $a_N$. Then, the dialogue semantic and style vectors of CD are matched with the stored dialogue semantic and style vectors in SDSSD, retrieving the Top-K most similar dialogues based on combined semantic and style similarity. 2) To effectively utilize the multi-source knowledge from CD and SD, we design a multi-source style knowledge aggregator to integrate style knowledge from SD, style knowledge from CD, and the predicted $a_N$ style knowledge. For SD, we use a Multi-granularity Heterogeneous Graph (MgHG) to model the dialogues retrieved from SD. First, we construct both text and audio MgHG structures for these dialogues, then apply a text MgHG encoder and a audio MgHG encoder to encode the text MgHG and audio MgHG, respectively, obtaining enhanced dialogue semantic and style features. Contrastive learning is then applied separately to the semantic and style features, pulling closer the features of positive examples while pushing away the representations of positive and negative examples. For CD, we also adopt the MgHG modeling mechanism to effectively capture and encode the semantic and style features of the CD. Finally, this knowledge is efficiently integrated through knowledge aggregation. 3) We feed the aggregated multi-source style knowledge into the speech synthesizer to help the agent synthesize expressive speech that aligns with the current conversational style. In summary, the main contributions of this paper include:

\begin{itemize}
    \item We propose a novel Retrieval-Augmented Dialogue Knowledge Aggregation CSS model, termed RADKA-CSS. To our knowledge, this is the first work to apply Retrieval-Augmented Generation to CSS.

    \item To facilitate the retrieval of dialogues similar to the current dialogue in terms of scenario and style, we build a stored dialogue semantic-style database based on DailyTalk and design a multi-attribute retrieval module. This module effectively retrieves similar dialogues by simultaneously considering both the semantics and style of dialogues.

    \item To better encode the retrieved dialogues and the current dialogue, we propose a multi-granularity heterogeneous graph modeling mechanism. This heterogeneous structure includes three types of nodes (word-level, sentence-level, and dialogue-level) and four types of relationships. It aims to capture the structural and temporal relations of nodes at different granularities to effectively encode semantic and style information.

    \item  Objective and subjective experiments show that the speech synthesized by RADKA-CSS outperforms all baseline models in terms of alignment with conversational style and expressiveness.
\end{itemize}

% 1) We propose a novel Retrieval-Augmented Dialogue Knowledge Aggregation CSS model, termed RADKA-CSS. To our knowledge, this is the first work to apply Retrieval-Augmented Generation to CSS.

% 2) To facilitate the retrieval of dialogues similar to the current dialogue in terms of scenario and style, we build a stored dialogue semantic-style database based on DailyTalk and design a multi-attribute retrieval module. This module effectively retrieves similar dialogues by simultaneously considering both the semantics and style of dialogues.

% 3) To better encode the retrieved dialogues and the current dialogue, we propose a multi-granularity heterogeneous graph modeling mechanism. This heterogeneous structure includes three types of nodes (word-level, sentence-level, and dialogue-level) and four types of relationships. It aims to capture the structural and temporal relations of nodes at different granularities to effectively encode semantic and style information.

% 4) Objective and subjective experiments show that the speech synthesized by RADKA-CSS outperforms all baseline models in terms of alignment with conversational style and expressiveness.

% \end{itemize}
 
The remainder of this paper is organized as follows: In Section \ref{sec2}, we briefly review some related works. In Section \ref{sec3}, we provide the task definition. In Section \ref{sec4}, we propose a novel RADKA-CSS framework. In Section \ref{sec5}, we introduce the experimental datasets and setups in detail. In Section \ref{sec6}, we conduct experiments to verify the effectiveness of RADKA-CSS. Finally, we conclude this paper and discuss future work in Section \ref{sec7}.

\section{Related Works}
\label{sec2}

\subsection{Retrieval-Augmented Generation in Dialogue}
Retrieval-augmented generation (RAG) endows pre-trained generative models with the ability to incorporate non-parametric memory, enabling them to utilize external knowledge effectively \cite{lewis2020retrieval}. In recent research, some works use RAG to handle dialogue tasks. For example, DFA-RAG \cite{sun2024dfarag} models dialogue as a deterministic finite automaton, guiding the dialogue through predefined states and retrieving the most relevant historical examples to generate contextually appropriate responses. UniMS-RAG \cite{wang2024unims} leverages multi-source knowledge retrieval, selecting and retrieving relevant information based on the dialogue content, and then generating personalized responses that align with the current dialogue context. ConvRAG \cite{ye2024boosting} proposes a conversation-level retrieval-augmented generation method. It refines the conversational question through interdependent dialogue history to better understand the question, uses a fine-grained retriever to obtain the most relevant information from the web, and applies a self-check mechanism to generate accurate responses. 
However, our RAG method has some clear differences from these works:  1) Multi-attribute retrieval enhancement: We first extract dialogue-level semantic and style vectors from the CD. The dialogue semantic and style vectors of CD are then matched with the dialogue semantic and style vectors in SDSSD based on similarity, and the Top-K most similar dialogues are retrieved based on the combined semantic and style similarity.  2) Expressive Conversational Speech Synthesis: While most RAG methods focus on generating text-based responses, our approach is specifically designed for CSS, effectively integrating retrieved knowledge to generate expressive speech that aligns with the conversational style.

\subsection{Contrastive Learning}
Contrastive learning pushes the embeddings of samples from different classes farther apart in the representation space while pulling the embeddings of samples from the same class closer together \cite{khosla2020supervised, le2020contrastive, tian2020makes, peng2023leader, liu2024contrastive, sun2024connecting}. For example, CALM \cite{meng22c_interspeech} optimizes the correlation between speaker style embeddings and style-related text features extracted from the text through contrastive learning, making the synthesized speech more expressive and natural. CLAPSpeech \cite{ye2023clapspeech} enhances prosody in synthesized speech by learning prosodic variations of the same text across different contexts through a cross-modal contrastive pre-training framework. CONCSS \cite{deng2024concss} enhances the discriminability of intra-modal context embeddings by incorporating contrastive learning constraints into the dialogue history context modeling. In our work, we employ contrastive learning to bring the semantic and style representations of similar dialogues closer. At the same time, we further distinguish the semantic and style representations of dialogues with different scenarios and inconsistent styles.

\subsection{Graph-Based Dialogue Modeling}
Graph Neural Network (GNN) is a type of neural network model designed to process graph-structured data \cite{sanchez-lengeling2021a}. It captures the relationships and structural information between nodes in a graph by learning representations of the nodes and edges \cite{sanchez-lengeling2021a, hamilton2017representation, hamilton2020graph, peng2022control}. 
% Previous GNN-based CSS methods use GNN models to understand dialogue context.
\cite{li2022enhancing} models both inter-speaker and intra-speaker dependencies within the dialogue using dialogue GNN, enhancing the speaking style of the synthesized speech. \cite{li2022inferring} infers speaking style from multimodal dialogue context using a multi-scale relational GNN. \cite{liu2024emotion} introduces heterogeneous graph modeling of multi-source knowledge to predict the emotion of the current sentence, synthesizing emotionally expressive speech. Our MgHG differs significantly from previous work in the following ways: 1) We employ three levels of granularity for nodes (word-level, sentence-level, and dialogue-level) to represent a set of dialogues. This approach of extracting features from local to global allows the model to more comprehensively capture the semantic and style features of the dialogue. 2) In constructing node relationships, we introduce parent-child relationships (words belong to sentence, sentences belong to dialogue) and sibling relationships (adjacent words in the same sentence, adjacent sentences in the same dialogue) to describe the relationships between multi-granularity nodes in the dialogue. This node connection approach enables the model to more effectively capture the granularity structure and temporal relationships within the dialogue, thereby enhancing the model's ability to understand and model the dialogue's semantic and style.

\begin{figure*}[ht]

    \centering    \includegraphics[width=1\textwidth]{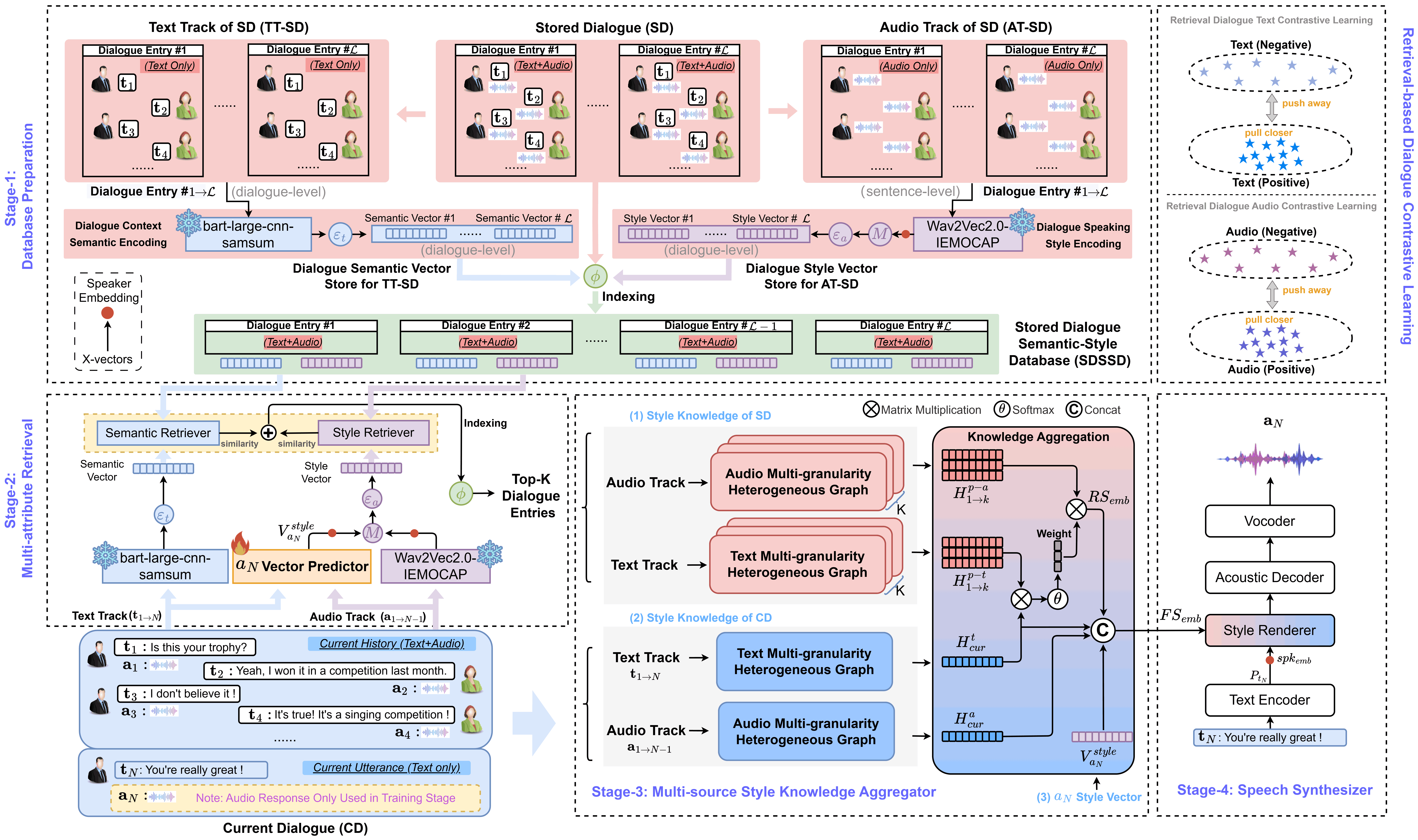}
    \caption{The overview of RADKA-CSS consists of Database Preparation, Multi-attribute Retrieval, Multi-source Style Knowledge Aggregator, and Speech Synthesizer.
} \vspace{-3mm}
    \label{ReKACSS-main}
\end{figure*}

\section{Task Definition}
\label{sec3}
% introduction  CD, SD, TT-SD, SDSSD.
Current dialogue can be defined as a sequence of utterances $\text{CD} = \{[t_1, a_1], ..., [t_{N-1}, a_{N-1}], [t_N, a_N]\}$, where $\{t_1, t_2, ..., t_{N-1}\}$ represents the text of the CD history and $t_N$ represents the text of the current sentence. $\{a_1, a_2, ..., a_{N-1}\}$ represents the audio of the CD history, and $a_N$ represents the speech to be synthesized. Stored Dialogue can be defined as  $\text{SD} = \{\text{Dialogue Entry \#1}, ..., \text{Dialogue Entry \#} \mathcal{L} \}$, where $\mathcal{L}$ represents the total number of dialogue entries contained in SD, and each dialogue entry represents a set of dialogues, including both text and audio modalities. We use the Text Track of SD (TT-SD) to represent SD which contains only the text modality, and the Audio Track of SD (AT-SD) to represent SD which contains only the audio modality. The reason for applying Retrieval-Augmented Generation to CSS is that SD comprises dialogue fragments from earlier phases of user-agent interactions, which include rich knowledge of conversational style expression. By integrating this knowledge, the agent can better synthesize speech that aligns with the current conversational style. Retrieval-augmented generation-based CSS needs to consider the following points: 1) How to design a retrieval scheme to ensure that the dialogues retrieved from SD are indeed similar in both scenario and style to the CD. 2) How to effectively model the CD and the retrieved dialogue content from SD to fully capture the conversational style features. 3) How to effectively apply the extracted conversational style features to the speech to be synthesized to generate expressive speech that aligns with the current conversational style.

\section{Methodology}
\label{sec4}
\subsection{Model Overview}
As shown in Fig. \ref{ReKACSS-main}, the proposed RADKA-CSS consists of four components: 1) Database Preparation, 2) Multi-attribute Retrieval, 3) Multi-source Style Knowledge Aggregator,  4) Speech Synthesizer. 
The database preparation aims to build a database, SDSSD, for dialogue semantic and style retrieval.
The multi-attribute retrieval extracts dialogue semantic and style vectors from CD and retrieves dialogues with similar scenarios and styles from SDSSD.
The multi-source style knowledge aggregator aims to encode style knowledge from both CD and the dialogues retrieved from SDSSD, and aggregates multi-source style knowledge.
The speech synthesizer uses integrated style knowledge to help the agent generate expressive speech that is aligned with the current conversational style.

\subsection{Database Preparation}
The upper left portion of Fig. \ref{ReKACSS-main} illustrates the process of database preparation. In building a database to support retrieval of dialogues similar to the current dialogue in terms of scenarios and style, we face two main challenges: 1) how to enable dialogue-level retrieval, and 2) how to effectively represent the scenario and style of a set of dialogues. To address these issues, we design two encoding modules: Dialogue Context Semantic Encoding and Dialogue Speaking Style Encoding. These modules extract dialogue-level semantic vectors and style vectors respectively from the text and audio features of a set of dialogues to capture the scenario and style of the dialogue.

\begin{figure}
    \centering
    \includegraphics[width=1\linewidth]{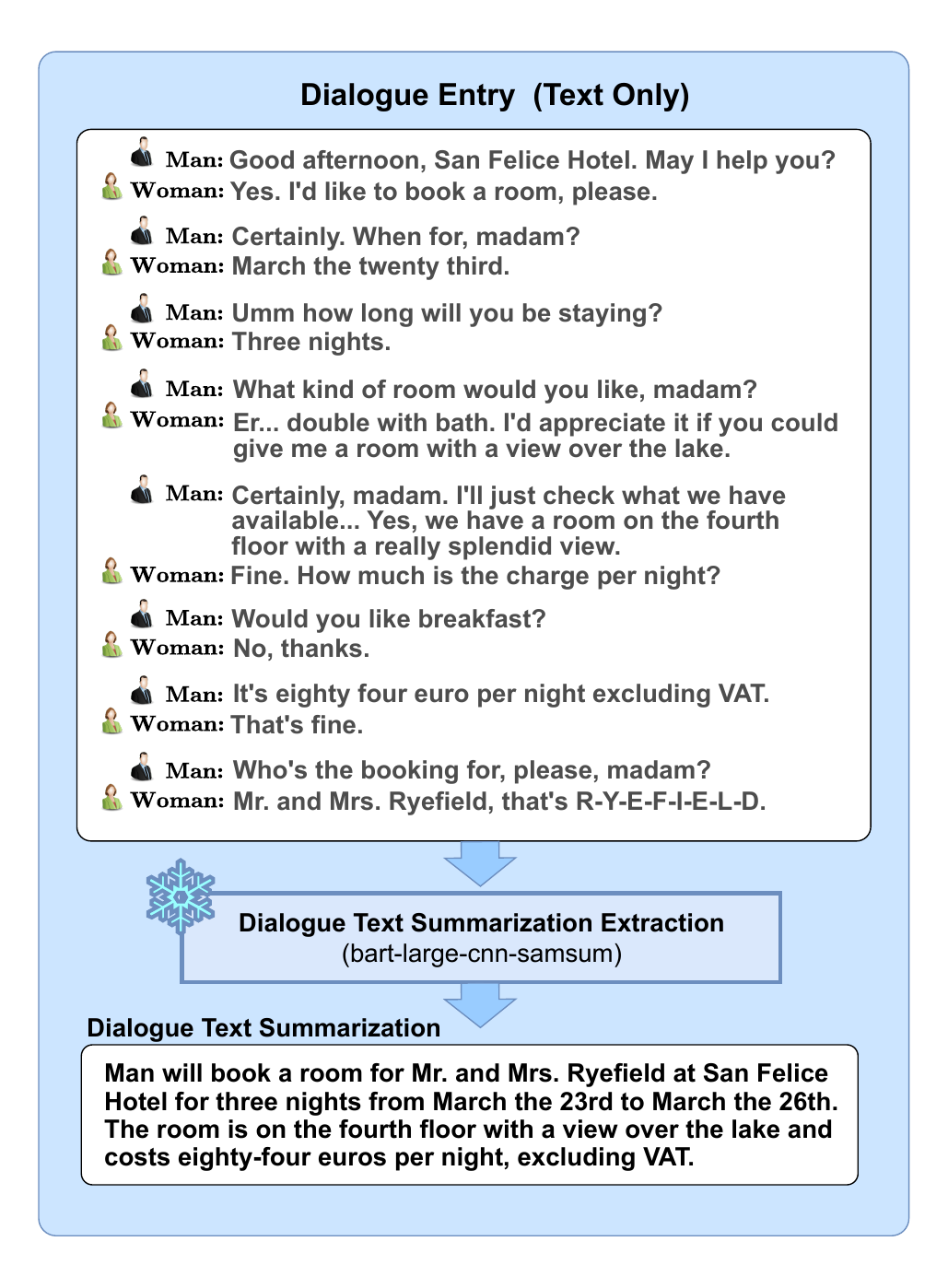}
    \vspace{-3mm}
    \caption{Example of Dialogue Text Summarization Extraction.}
    \vspace{-3mm}
    \label{summary}
\end{figure}

Specifically, SD contains $\mathcal{L}$ sets of dialogue entries with man and woman speakers alternating, covering both text and audio modalities. We first separate these modalities into TT-SD, which includes only text, and AT-SD, which includes only audio. We combine text dialogue entries from TT-SD into a long paragraph and input it into the Dialogue Text Summarization Extraction (bart-large-cnn-samsum\footnote{\label{samsum}https://huggingface.co/philschmid/bart-large-cnn-samsum}) to generate a dialogue text summarization with speaker information, as shown in Fig. \ref{summary}. Next, we use Sentence-BERT\footnote{\label{sentencebert}https://huggingface.co/sentence-transformers/distiluse-base-multilingual-cased-v1} \cite{reimers-gurevych-2019-sentence} to vectorize the summarization, obtaining semantic vectors to represent dialogue scenarios. For dialogue style, we input audio dialogue entries from AT-SD into Wav2Vec2.0-IEMOCAP\footnote{\label{wav2vec2iemocap}https://huggingface.co/speechbrain/emotion-recognition-wav2vec2-IEMOCAP} sentence by sentence, generating sentence-level style vectors. We then use X-vectors\footnote{\label{x-vector}https://huggingface.co/speechbrain/spkrec-xvect-voxceleb} \cite{xvector} to encode speaker information, adding speaker information to these style vectors. Finally, we aggregate the sentence-level style vector of each dialogue entry to represent the overall style characteristics of the dialogue.

In conclusion, we build the stored dialogue semantic-style database by combining dialogue entries with text and audio, along with their corresponding dialogue semantic and style vectors.

\subsection{Multi-attribute Retrieval}
To retrieve dialogues from SDSSD that are similar to the CD in terms of scenario and style, we design a multi-attribute retrieval module that combines both semantic and style attributes, as shown in the bottom left part of Fig. \ref{ReKACSS-main}.

For semantic attribute retrieval, we aim to retrieve dialogues with scenarios similar to the CD. We use the same method as in building SDSSD to obtain the dialogue’s semantic vector. Specifically, we merge the text of the CD, $t_{1 \rightarrow N}$, into a long paragraph, as shown in Fig. \ref{summary}, extract the dialogue text summarization using the dialogue text summarization extraction (bart-large-cnn-samsum\footref{samsum}), and then vectorize it through Sentence-BERT\footref{sentencebert} to generate the semantic vector of the CD. This semantic vector is then input into the semantic retriever along with the dialogue semantic vectors in SDSSD for similarity calculation.

For style attribute retrieval, we aim to retrieve dialogues with styles similar to the CD. Specifically, the absence of the audio $a_N$ leaves the conversational style incomplete. To address this issue, we design an $a_N$ vector predictor to predict the sentence-level style vector of $a_N$ ($V^{style}_{a_N}$). The $a_N$ vector predictor includes a text context encoder and an audio context encoder. Both the text context encoder and the audio context encoder consist of one layer of bidirectional GRU and two layers of linear layers. In addition, we input $a_{1 \rightarrow N-1}$ into Wav2Vec2.0-IEMOCAP to extract sentence-level style vectors. We then add the speaker information extracted from the X-vectors to these sentence-level style vectors and aggregate them to generate the style vector of the CD. This style vector is subsequently input into the style retriever, along with the dialogue style vectors in SDSSD, for similarity calculation.

Finally, to integrate the results of semantic and style attribute retrieval, we add the similarity scores obtained from dialogue semantics and style. Based on this combined similarity, we select the Top-K dialogue entries.

\subsection{Multi-source Style Knowledge Aggregator}
The multi-source style knowledge aggregator aims to encode style knowledge from both CD and the dialogues retrieved from SD and aggregate the style knowledge of SD, style knowledge of CD, and the predicted $a_N$ style vector ($V^{style}_{a_N}$).

\subsubsection{Style Knowledge of SD}
To better encode the style and semantic information of the Top-K dialogue entries, we design the Audio Multi-granularity Heterogeneous Graph (AMgHG) and Text Multi-granularity Heterogeneous Graph (TMgHG), which encode the audio track and text track of the Top-K dialogue entries, respectively.

For AMgHG, consists of two parts: 1) \textbf{AMgHG Initialization}, which constructs the audio track of the Top-K dialogue entries into a heterogeneous graph structure with three levels of granularity and four types of relationships, and initializes the features of word-level, sentence-level, and dialogue-level nodes for audio. 2) \textbf{AMgHG Encoder}, which encodes the AMgHG to fully capture the style representations.

\textbf{AMgHG Initialization. }
Inspired by real-world conversations, understanding a complete set of dialogues typically requires consideration from three angles: 1) the theme of the entire dialogue (dialogue-level), 2) the contribution of each sentence to the overall conversational style (sentence-level), and 3) the impact of each word on the meaning of the sentence (word-level). Therefore, when constructing the AMgHG, we establish three levels of nodes: word-level, sentence-level, and dialogue-level. This allows for a gradual transition from local features to global features, enabling the model to capture the style information of the dialogue more comprehensively. These nodes include four types of relationships: parent-child relationships (words belong to sentence, sentences belong to dialogue) and sibling relationships (adjacent words in the same sentence, adjacent sentences in the same dialogue). These four relationships allow the nodes of the AMgHG to fully aggregate contextual style information from both the temporal relationships and the granularity structure. For the k-th dialogue entry, We extract the dialogue-level style node ($R^{a_k}_{d}$), sentence-level style nodes ($R^{a_k}_{s_{1 \rightarrow N}}$), and word-level style nodes ($R^{a_k}_{w_{1 \rightarrow N, 1 \rightarrow q}}$).

\begin{itemize}
    \item \textbf{Dialogue-level style node:} We first use Wav2Vec2.0-IEMOCAP\footref{wav2vec2iemocap} to extract sentence-level style features, then apply Average Pooling to aggregate the dialogue-level style node: $R^{a_k}_{d}$.

    \item \textbf{Sentence-level style nodes:} We use Wav2Vec2.0-IEMOCAP\footref{wav2vec2iemocap} to extract $R^{a_k}_{s_{1 \rightarrow N}}$

    \item \textbf{Word-level style nodes:} We utilize MFA to identify the start and end positions of each word in the audio, then use Wav2Vec2.0\footnote{\label{wav2vec2}https://huggingface.co/facebook/wav2vec2-base-960h} to extract frame-level style features and apply Average Pooling to obtain $R^{a_k}_{w_{1 \rightarrow N, 1 \rightarrow q}}$.
\end{itemize}

\begin{figure}
    \centering
    \vspace{-3mm}\includegraphics[width=1\linewidth]{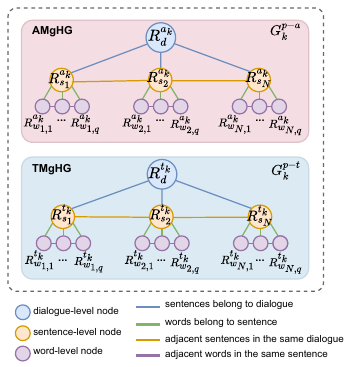}
    \vspace{-4mm}
        \caption{The structural diagrams of AMgHG and TMgHG.}
    \vspace{-6mm}
    \label{graph}
\end{figure}

 As shown in Fig. \ref{graph}. these three levels of style nodes form an AMgHG ($G^{p-a}_{k}$).

\textbf{AMgHG Encoder. }
For a given k-th AMgHG $G^{p-a}_{k}$, each node aggregates information from its neighboring nodes through different relationships.  Specifically, the sentence-level style node $R^{a_k}_{s_i}$ aggregates information from the dialogue-level style node $R^{a_k}_d$, the neighboring sentence-level style nodes $R^{a_k}_{s_{i-1}}$ and $R^{a_k}_{s_{i+1}}$, as well as the word-level style nodes $R^{a_k}_{w_{i, 1 \rightarrow q}}$. Once all node information is aggregated, the three granularity nodes are fused. Firstly, $R^{a_k}_{s_{1 \rightarrow N}}$ and $R^{a_k}_{w_{1 \rightarrow N, 1 \rightarrow q}}$ are fed into bidirectional LSTM separately to fuse the sentence-level and word-level style node, outputting sentence-level contextual style features $R^{a_k}_{s-agg}$ and word-level contextual style features $R^{a_k}_{w-agg}$. Finally, $R^{a_k}_{s-agg}$, $R^{a_k}_{w-agg}$, and $R^{a_k}_d$ are concatenated along the feature dimension, and then fused through a linear layer to represent the style features of the k-th dialogue: $H^{p-a}_ {k}$.

For TMgHG, consists of two parts: 1) \textbf{TMgHG Initialization}, which constructs the text track of the Top-K dialogue entries into a heterogeneous graph structure with three levels of granularity and four types of relationships, and initializes the features of word-level, sentence-level, and dialogue-level nodes for text. 2) \textbf{TMgHG Encoder}, which encodes the TMgHG to fully capture the semantic representations.

\textbf{TMgHG Initialization. }
The structure of TMgHG is the same as that of AMgHG. For the k-th dialogue entry, 
We extract the dialogue-level semantic node ($R^{t_k}_{d}$), sentence-level semantic nodes ($R^{t_k}_{s_{1 \rightarrow N}}$), and word-level semantic nodes ($R^{t_k}_{w_{1 \rightarrow N, 1 \rightarrow q}}$).

\begin{itemize}
    \item \textbf{Dialogue-level semantic node:} We first use bart-large-cnn-samsum\footref{samsum} to extract the summarization of the dialogue text, and then the summarization is fed into Sentence-BERT\footref{sentencebert} to extract $R^{t_k}_{d}$.

    \item \textbf{Sentence-level semantic nodes:} We input each sentence from the dialogue into Sentence-BERT\footref{sentencebert} to extract $R^{t_k}_{s_{1 \rightarrow N}}$.

    \item \textbf{Word-level semantic nodes:} We use TOD-BERT\footnote{\label{tod-bert}https://huggingface.co/TODBERT/TOD-BERT-JNT-V1} to extract $R^{t_k}_{w_{1 \rightarrow N, 1 \rightarrow q}}$.
\end{itemize}

 As shown in Fig. \ref{graph}. these three levels of semantic nodes form a TMgHG ($G^{p-t}_{k}$).

\textbf{TMgHG Encoder. }
For a given k-th TMgHG $G^{p-t}_{k}$, each node aggregates information from its neighboring nodes through different relationships.  Specifically, the sentence-level semantic node $R^{t_k}_{s_i}$ aggregates information from the dialogue-level semantic node $R^{t_k}_d$, the neighboring sentence-level semantic nodes $R^{t_k}_{s_{i-1}}$ and $R^{t_k}_{s_{i+1}}$, as well as the word-level semantic nodes $R^{t_k}_{w_{i, 1 \rightarrow q}}$. Once all node information is aggregated, the three granularity nodes are fused. Firstly, $R^{t_k}_{s_{1 \rightarrow N}}$ and $R^{t_k}_{w_{1 \rightarrow N, 1 \rightarrow q}}$ are fed into bidirectional LSTM separately to fuse the sentence-level and word-level semantic node, outputting sentence-level contextual semantic features $R^{t_k}_{s-agg}$ and word-level contextual semantic features $R^{t_k}_{w-agg}$. 
Finally, $R^{t_k}_{s-agg}$, $R^{t_k}_{w-agg}$, and $R^{t_k}_d$ are concatenated along the feature dimension, and then fused through a linear layer to represent the semantic features of the k-th dialogue: $H^{p-t}_{k}$.

\subsubsection{Style Knowledge of CD}
For CD, we adopt AMgHG and TMgHG structures identical to those used for processing SD, with shared parameters, aiming to capture and encode the style and semantic information of the current dialogue. Specifically, we input $a_{1 \rightarrow N-1}$ and $t_{1 \rightarrow N}$ into AMgHG Initialization and TMgHG Initialization, respectively, to obtain $G^{t}_{cur}$ and $G^{a}_{cur}$. These are then encoded by AMgHG Encoder and TMgHG Encoder to derive the style feature $H^{a}_{cur}$ and semantic feature $H^{t}_{cur}$ for the CD.

\subsubsection{Knowledge Aggregation}
To effectively aggregate the style knowledge of SD, style knowledge of CD, and the predicted $a_N$ style vector ($V^{style}_{a_N}$), we design a knowledge aggregation method inspired by \cite{meng22c_interspeech}. 
\vspace{-3mm}
\begin{equation}
    \begin{aligned}
    &W=Softmax(H^{p-t}_{1 \rightarrow k} \cdot H^{t}_{cur})
        \\&RS_{emb}=W^{T}\cdot H^{p-a}_{1 \rightarrow k}
        \\&FS_{emb}= Concat(RS_{emb}, H^{t}_{cur}, H^{a}_{cur}, V^{style}_{a_N} )
    \end{aligned}
    \label{equation4}
\end{equation}

As shown in Equation \ref{equation4}, we first calculate the weights $W$ using $H^{p-t}_{1 \rightarrow k}$ and $H^{t}_{cur}$. Then, based on the weights $W$, we fuse $H^{p-a}_{1 \rightarrow k}$ to obtain the retrieved style embedding ($RS_{emb}$). Finally, we concatenate $RS_{emb}$, $H^{t}_{cur}$, $H^{a}_{cur}$, and $V^{style}_{a_N}$ to obtain the final style embedding $FS_{emb}$.

\subsection{Retrieval-based Dialogue Contrastive Learning}
We consider that dialogues similar to CD (positive samples) are close to CD in both scenario and style. In contrast, dialogues dissimilar to CD (negative samples) are not only inconsistent with CD in terms of both dialogue scenario and style but also have different scenarios and styles among themselves. Therefore, we design a Retrieval-based Dialogue Contrastive Learning module, as shown in the upper right corner of Fig. \ref{ReKACSS-main}. This module includes Retrieval Dialogue Text Contrastive Learning ($\mathcal{L}^{cl}_{t}$) and Retrieval Dialogue Audio Contrastive Learning ($\mathcal{L}^{cl}_{a}$). $\mathcal{L}^{cl}_{t}$ aims to pull the semantic representations of positive samples closer while pushing away the semantic representations between positive and negative samples. $\mathcal{L}^{cl}_{a}$ aims to pull the style representations of positive samples closer while pushing away the style representations between positive and negative samples. For positive dialogue semantic representations $H^{p-t} = [H^{p-t}_{1}, ..., H^{p-t}_{k}]$, and negative dialogue semantic representations $H^{n-t} = [H^{n-t}_{1}, ..., H^{n-t}_{k}]$, $H^{pn-t}$ represents $H^{p-t} \cup H^{n-t}$, $\mathcal{L}^{cl}_{t}$ for $H^{p-t}_{k}$ is as follows:

\begin{equation}
    \mathcal{L}^{cl}_{t}= -log \frac{\underset{H^{p-t}_{i}\in H^{p-t}}{\sum } exp(sim(H^{p-t}_{k}, H^{p-t}_{i})/\tau)}{\underset{H^{pn-t}_{j}\in H^{pn-t}}{\sum } exp(sim(H^{p-t}_{k}, H^{pn-t}_{j})/\tau)}
\end{equation}
where $sim(\cdot,\cdot)$ is a cosine similarity function. $\tau$ is a scalar temperature parameter.

For positive dialogue style representations $H^{p-a} = [H^{p-a}_{1}, ..., H^{p-a}_{k}]$, and negative dialogue style representations $H^{n-a} = [H^{n-a}_{1}, ..., H^{n-a}_{k}]$, $H^{pn-a}$ represents $H^{p-a} \cup H^{n-a}$, $\mathcal{L}^{cl}_{a}$ for $\mathcal{H}^{p-a}_{k}$ is as follows:

\begin{equation}
    \mathcal{L}^{cl}_{a}= -log \frac{\underset{H^{p-a}_{i}\in H^{p-a}}{\sum } exp(sim(H^{p-a}_{k}, H^{p-a}_{i})/\tau)}{\underset{H^{pn-a}_{j}\in H^{pn-a}}{\sum } exp(sim(H^{p-a}_{k}, H^{pn-a}_{j})/\tau)}
\end{equation}
where $sim(\cdot,\cdot)$ is a cosine similarity function. $\tau$ is a scalar temperature parameter.

\subsection{Speech Synthesizer}
As shown in Fig. \ref{ReKACSS-main}, the speech synthesizer is based on FastSpeech2 \cite{ren2021fastspeech}. It consists of a text encoder, an acoustic decoder, and a vocoder. The text encoder aims to extract phoneme-level linguistic encodings $P_{t_N}$ for the current utterance. The acoustic decoder includes a length regulator and a variance adapter to predict duration, energy, and pitch, followed by a Mel decoder to predict Mel spectrogram features. Finally, a pre-trained HiFi-GAN \cite{kong2020hifi} vocoder is used to generate speech. Note that we add a new Style Renderer on top of the text encoder to add style knowledge into the linguistic encodings by a set of trainable weight parameters. Specifically, we take $FS_{emb}$ from the Multi-source Style Knowledge Aggregator, $P_{t_N}$ from the TTS Encoder, and the current speaker id embedding ($spk_{emb}$) as inputs, integrating the conversational style into the phoneme-level speech encoding to synthesize expressive speech that aligns with the current conversational style.

\section{Experiments}
\label{sec5}
\subsection{Dataset}
We validate RADKA-CSS on the DailyTalk \cite{lee2023dailytalk} dataset for CSS. 
DailyTalk is a benchmark dataset that is widely used in many related works \cite{deng2023cmcu, xue2023m,liu2024emotion, witzig2024multimodal}, and we believe it is enough and very convincing to conduct experiments on this dataset. In addition, more importantly, DailyTalk is a high-quality recording data with only 2 speakers, which is most suitable for database construction and RAG implementation. In other multi-speaker data, the amount of data from different speakers varies too much to be used for database construction and retrieval.

DailyTalk consists of 23,773 audio clips, representing 20 hours and 2,541 conversations. Each conversation contains an average of 9.356 turns, with an average clip length of 3.282 seconds. The dataset is recorded simultaneously by a male and a female speaker, with a balanced number of utterances from each. All speech samples are recorded at a sampling rate of 44.10 kHz and encoded in 16-bit format. We partition the data into training, validation, and test sets in an 8:1:1 ratio.

\subsection{Implementation Details}
The $a_N$ vector Predictor includes a text context encoder and an audio context encoder. Both the text context encoder and the audio context encoder consist of one layer of bidirectional GRU and two layers of linear layers. The structures of the Text Multi-granularity Heterogeneous Graph and Audio Multi-granularity Heterogeneous Graph are the same: First, the HeteroConv layer, composed of four SAGEConv layers, in channels and out channels set to 256. Next, three linear layers project the dialogue-level, sentence-level, and word-level nodes into an embedding space with an output dimension of 256. Following this, two bidirectional LSTM, with input and output dimensions of 256, fuse the sentence-level and word-level node features. Finally, the dialogue-level node feature, the fused sentence-level node features, and the word-level node features are concatenated and fed into a linear layer with an input dimension of 768 and an output dimension of 256. We employ the Adam optimizer for speech synthesizer with $\beta$1 = 0.9 and $\beta$2 = 0.98. Grapheme-to-Phoneme (G2P\footnote{\label{g2p}https://github.com/Kyubyong/g2p}) toolkit is used for converting all text inputs into their respective phoneme sequences. We utilize the Montreal Forced Alignment (MFA) \cite{mcauliffe17_interspeech} tool to extract phoneme duration alignment. All speech samples are re-sampled to 22.05 kHz. Mel-spectrum features are extracted with a window length of 25ms and a shift of 10ms. The model is trained on an A800 GPU with a batch size of 16. Model optimization \cite{werbos1990backpropagation, jia2020design, xiao2020finite, yuan2020exponential} achieves optimal performance at 300k steps.

% ---------------------------------------
\begin{table*}[h]
\centering

\caption{N-DMOS and S-DMOS subjective evaluation scoring criteria.}
\label{dmos}

\begin{tabular}{c|c|c}
\hline
\textbf{Scale} & \textbf{N-DMOS} &  \textbf{S-DMOS} \\
\hline
5  & Excellent speech quality and naturalness & Perfectly matches the current conversational style \\
4  & Good speech quality and naturalness & Consistent with the current conversational style \\
3  & Fair speech quality and naturalness & Moderately matches the current conversational style \\
2  & Poor speech quality and naturalness & Slightly mismatches the current conversational style \\
1  & Very poor speech quality and naturalness & Completely mismatches the current conversational style \\
\hline
\end{tabular}

\end{table*}
% ---------------------------------------

\begin{table*}[h]
\centering

\caption{N-CMOS and S-CMOS subjective evaluation scoring criteria.}
\label{cmos}

\begin{tabular}{c|c|c}
\hline
\textbf{Scale} & \textbf{N-CMOS (A vs. B Comparison)} &  \textbf{S-CMOS (A vs. B Comparison)} \\
\hline
3  & A is much more natural than B & A aligns significantly better with the conversational style than B \\
2  &  A is more natural than B & A aligns better with the conversational style than B \\
1  & A is slightly more natural than B & A aligns slightly better with the conversational style than B \\
0  & Both are about the same in naturalness & Both are equally aligned with the conversational style \\
-1  & B is slightly more natural than A & B aligns slightly better with the conversational style than A \\
-2  & B is more natural than A & B aligns better with the conversational style than A \\
-3  & B is much more natural than A & B aligns significantly better with the conversational style than A \\
\hline
\end{tabular}

\end{table*}

% ---------------------------------------

\subsection{Comparative Models}
RADKA-CSS requires the retrieval of dialogues with similar scenarios and styles to help the agent synthesize speech with a conversational style. Although the dialogue RAG retrieval methods provided by related works, such as DFA-RAG \cite{sun2024dfarag}, UniMS-RAG \cite{wang2024unims}, and ConvRAG \cite{ye2024boosting}, are mentioned, they do not meet our specific retrieval requirements. Therefore, we did not include these models as comparative models. To demonstrate the effectiveness of our RADKA-CSS, we compare it against five state-of-the-art CSS models, all utilizing FastSpeech2 as the TTS backbone.

\begin{itemize}

\item \textbf{DailyTalk} \cite{lee2023dailytalk} incorporates a dialogue context encoder based on \cite{guo2021conversational} into FastSpeech2 \cite{ren2021fastspeech} to model sentence-level text dialogue history.

\item \textbf{M$^2$-CTTS} \cite{xue2023m} designs both coarse-grained and fine-grained text and speech context modeling modules, aiming to fully leverage multimodal history to enhance the prosodic expression in synthesized speech.

\item \textbf{Homogeneous Graph-based CSS} \cite{li2022inferring} proposes a context modeling method based on a multi-scale relational graph convolutional network, which models both global and local dependencies in multimodal dialogue history to enhance its ability to synthesize speaking style.

\item \textbf{CONCSS} \cite{deng2024concss} introduces a CSS framework based on contrastive learning, which incorporates a negative-sample-enhanced sampling strategy to improve the discriminability of context vectors, resulting in synthesized speech with better context adaptation and prosody sensitivity.

\item \textbf{ECSS} \cite{liu2024emotion} presents a new emotion CSS model based on heterogeneous graph-based emotion context modeling and an emotion rendering mechanism, ensuring the accurate generation of emotional conversational speech in terms of both emotion understanding and expression.

\end{itemize}

\subsection{Ablation Models}
We conduct thorough ablation experiments to validate the contributions of different components of RADKA-CSS. The details are as follows:
\begin{itemize}
    \item \textbf{Abl.1: w/o Style Knowledge (SD)} indicates the removal of similar dialogue knowledge retrieved from SD, aiming to verify whether referencing the style knowledge of similar dialogues in SD enhances the agent's ability to understand the style of the CD.

    \item \textbf{Abl.2: w/o Style Knowledge (CD: Text)} indicates that encoded textual knowledge from the CD is not aggregated, aiming to validate the extent to which the semantic information contained in the scenarios of CD contributes to the agent's ability to generate speech that aligns with the conversational style.

    \item \textbf{Abl.3: w/o Style Knowledge (CD: Audio)} indicates that encoded audio knowledge from the CD is not aggregated, which assesses the effectiveness of the conversational style knowledge in CD and its role in enhancing the agent's understanding of the CD style.

    \item \textbf{Abl.4: w/o Style Knowledge ($a_N$ Style Vector)} indicates that the $a_N$ style knowledge predicted by the $a_N$ vector predictor is not aggregated, aiming to evaluate whether the predicted $a_N$ style vector contains style information.
    
    \item \textbf{Abl.5: w/o Heterogeneous Graph} replaces the heterogeneous graph in RADKA-CSS with a homogeneous graph to validate whether the proposed heterogeneous structure can better capture and model dialogue style and semantic representations.
    
    \item \textbf{Abl.6: w/o Multi-granularity} removes the dialogue-level and word-level nodes, using only sentence-level nodes for training, to evaluate whether the multi-granularity approach can more comprehensively represent style and semantic features.
    
    \item \textbf{Abl.7: w/o Knowledge Aggregation Method} replaces the knowledge aggregation method with simple direct addition, which helps validate the effectiveness of our proposed style knowledge aggregation method and its impact on performance.

    \item \textbf{Abl.8: w/o Contrastive Learning} removes the retrieval-based dialogue contrastive learning to validate its effectiveness and its impact on RADKA-CSS's performance.
    
    \item \textbf{Abl.9: w/ GT (Retrieved)} uses the ground truth Top-K dialogue entry instead of the Top-K dialogue entry retrieved during inference. This tests whether using Top-K dialogues that are more similar to the CD can better help RADKA-CSS in generating speech that aligns with the current conversational style.

\end{itemize}

\subsection{Evaluation Metrics }
For subjective evaluation metrics, we organize a DMOS (Dialogue Mean Opinion Score) \cite{streijl2016mean}  and CMOS (Comparative Mean Opinion Score) \cite{guo2021conversational} listening test with 20 graduate students who speak English as a second language. 
All listeners are graduate students in the field of speech who have passed the CET-6, IELTS, or TOEFL exams. They have extensive experience in DMOS scoring and have received specialized training on DMOS evaluation guidelines. In the DMOS test, listeners rate the synthesized speech of the current utterance on naturalness DMOS (N-DMOS) and style DMOS (S-DMOS) using a scale from 1 to 5, based on the dialogue history. Note that N-DMOS and S-DMOS need to be tested separately on the same synthesized speech samples and under the same listener conditions. N-DMOS focuses on the quality and naturalness of the speech, while S-DMOS evaluates whether the expression of the current utterance aligns with the conversational style. For the CMOS test, we similarly ask listeners to first listen to the dialogue history and then rate the two comparative models based on the dialogue context, using a score range from -3 to 3, where N-CMOS is used to compare the naturalness of the synthesized speech, and S-CMOS is used to compare the consistency of the synthesized speech with the current conversation style. The criteria for DMOS are detailed in Table \ref{dmos}, and the criteria for CMOS are detailed in Table \ref{cmos}.

For objective evaluation metrics, we calculate the Mean Absolute Error (MAE) \cite{willmott2005advantages} between the predicted and ground-truth acoustic features to assess the style expressiveness of the synthesized speech. Specifically, we use MAE-P, MAE-E, and MAE-D to evaluate pitch, energy, and duration acoustic features.

To validate that the dialogues retrieved by RADKA-CSS align with the style and content of the current conversation, we use Recall \cite{zhu2004recall} to evaluate the performance of retrieving ground-truth Top-K dialogue indices during inference. Specifically, the ground-truth Top-K dialogues are selected by matching the dialogue semantic vectors and dialogue style vectors using a pre-trained model, and the K most similar dialogues are chosen. Then, volunteers re-listen to these dialogues, read their content, and reorder the K dialogues based on their similarity in both scenario and style to the current dialogue.
% \begin{equation}
% \text{Recall} = \frac{\text{TP}}{\text{TP} + \text{FN}}
% \end{equation}
\begin{equation}
\text{Recall} = \frac{\text{TP}}{\text{TP} + \text{FN}}
\end{equation}
where TP and FN represent the numbers of true positive and false negative samples, respectively.

% ----------------------------------------

\begin{table*}[t!]
\centering
\caption{Subjective (with 95\% confidence interval \cite{yasuda23_interspeech}) and objective results with different comparative models.}
\setlength{\tabcolsep}{4pt} 
% \small 
\begin{tabular}{lccccccccccccc}
\hline
\textbf{Systems}  &  &  &  &  &  &  &  & \textbf{N-DMOS} ($\uparrow$)  & \textbf{S-DMOS} ($\uparrow$)   &  & \textbf{MAE-P} ($\downarrow$)   & \textbf{MAE-E} ($\downarrow$)   & \textbf{MAE-D} ($\downarrow$)     \\ \hline

DailyTalk \cite{lee2023dailytalk}     &  &  &  &  &  &  &  & 3.453 $\pm$ 0.024    & 3.434 $\pm$ 0.024    &  & 0.530   & 0.467   & 0.204     \\

M$^2$-CTTS \cite{xue2023m}     &  &  &  &  &  &  &  & 3.551 $\pm$ 0.023    & 3.452 $\pm$ 0.021    &  & 0.543   & 0.380   & 0.146   \\

Homogeneous Graph-based CSS \cite{li2022inferring}    &  &  &  &  &  &  &  & 3.599 $\pm$ 0.018    & 3.511 $\pm$ 0.027    &  & 0.489   & 0.320   & 0.146   \\

CONCSS \cite{deng2024concss}    &  &  &  &  &  &  &  & 3.688 $\pm$ 0.015    & 3.647 $\pm$ 0.022    &  & 0.482   & 0.328   & 0.143   \\ 

ECSS \cite{liu2024emotion}     &  &  &  &  &  &  &  & 3.720 $\pm$ 0.023    & 3.698 $\pm$ 0.021    &  & 0.505   & 0.332   & 0.134   \\ \hline

\textbf{RADKA-CSS (Proposed)}  &  &  &  &  &  &  &  & \textbf{3.904 $\pm$ 0.022}   & \textbf{3.879 $\pm$ 0.025}   &  & \textbf{0.442}   & \textbf{0.305}  & \textbf{0.130}     \\ \hline

\textbf{Ground Truth}  &  &  &  &  &  &  &  & \textbf{4.448 $\pm$ 0.021}   & \textbf{4.498 $\pm$ 0.023}   &  & -   & -  & -   \\ \hline

\end{tabular}

\label{cexp}
\end{table*}

% ---------------------------------- 

\begin{table*}[t!]
\centering
\caption{Subjective (with 95\% confidence interval \cite{yasuda23_interspeech}) and objective results with different ablation models.}
\setlength{\tabcolsep}{4pt} 
% \small 
\begin{tabular}{lccccc}
\hline
\textbf{Systems}  & \textbf{N-DMOS} ($\uparrow$)  & \textbf{S-DMOS} ($\uparrow$)   & \textbf{MAE-P} ($\downarrow$)   & \textbf{MAE-E} ($\downarrow$)   & \textbf{MAE-D} ($\downarrow$)     \\ \hline

\textbf{RADKA-CSS (Proposed)}  & \textbf{3.904 $\pm$ 0.022}   & \textbf{3.879 $\pm$ 0.025}   & \textbf{0.442}   & \textbf{0.305}  & \textbf{0.130}     \\ \hline

\qquad Abl.1:  w/o Style Knowledge (SD)   & 3.707 $\pm$ 0.022    & 3.694 $\pm$ 0.021    & 0.463   & 0.323   & 0.139   \\

\qquad Abl.2:  w/o Style Knowledge (CD: Text)   & 3.769 $\pm$ 0.018    & 3.740 $\pm$ 0.022    & 0.453   & 0.315   & 0.132   \\

\qquad Abl.3:  w/o Style Knowledge (CD: Audio)   & 3.726 $\pm$ 0.021    & 3.701 $\pm$ 0.019    & 0.458   & 0.317   & 0.132   \\

\qquad Abl.4:  w/o Style Knowledge ($a_N$ Style Vector)   & 3.775 $\pm$ 0.022    & 3.751 $\pm$ 0.023    & 0.456   & 0.314   & 0.131   \\

\qquad Abl.5:  w/o Heterogeneous Graph   & 3.732 $\pm$ 0.019    & 3.724 $\pm$ 0.021    & 0.471   & 0.327   & 0.140   \\

\qquad Abl.6:  w/o Multi-granularity   & 3.685 $\pm$ 0.026    & 3.683 $\pm$ 0.023    & 0.473   & 0.322   & 0.142   \\

\qquad Abl.7:  w/o Knowledge Aggregation Method   & 3.754 $\pm$ 0.023    & 3.723 $\pm$ 0.024    & 0.465   & 0.311   & 0.133   \\

\qquad Abl.8:  w/o Contrastive Learning   & 3.757 $\pm$ 0.017    & 3.699 $\pm$ 0.019    & 0.458   & 0.318   & 0.131   \\

\qquad Abl.9:  w/ GT (Retrieved)   & 3.962 $\pm$ 0.021    & 3.906 $\pm$ 0.017    & 0.440   & 0.301   & 0.131   \\ \hline

\end{tabular}

\label{aexp}
\end{table*}

%  ------------------------------------

\section{Results and Discussion}
\label{sec6}

\subsection{Main Results}
% As shown in Table \ref{cexp}, the overall performance of RADKA-CSS reaches the optimal level. In terms of subjective metrics, RADKA-CSS outperforms all baseline models with N-DMOS (3.904) and S-DMOS (3.879). In terms of objective metrics, MAE-P (0.442), MAE-E (0.305), and MAE-D (0.130) all achieve the best results.  This reflects the advantages of our RADKA-CSS. RADKA-CSS better understands the current conversational style by combining the style knowledge retrieved from early dialogue fragments in SD with the style knowledge in CD, enabling it to generate speech that aligns with the current conversational style.

In this section, we analyze the comparison results of RADKA-CSS with five state-of-the-art CSS models. As shown in Table \ref{cexp}, comparing all baseline models except RADKA-CSS. DailyTalk \cite{lee2023dailytalk} shows the poorest results in subjective metrics N-DMOS (3.453) and S-DMOS (3.434), as well as objective metrics MAE-P (0.530), MAE-E (0.467), and MAE-D (0.204). In contrast, ECSS \cite{liu2024emotion} achieves the best results in N-DMOS (3.720), S-DMOS (3.698), and MAE-D (0.134). CONCSS \cite{deng2024concss} performs best in MAE-P (0.482), while Homogeneous Graph-based CSS \cite{li2022inferring} achieves the best result in MAE-E (0.320). Note that, compared to these best baseline models, our RADKA-CSS shows significant improvements. In subjective metrics, RADKA-CSS outperforms all comparative models with N-DMOS (3.904) and S-DMOS (3.879). In terms of objective metrics, it achieves the best results with MAE-P (0.442), MAE-E (0.305), and MAE-D (0.130). Specifically, RADKA-CSS improves by 0.184 in N-DMOS, 0.181 in S-DMOS, and 0.004 in MAE-D compared to ECSS, improves by 0.040 in MAE-P compared to CONCSS, and improves by 0.015 in MAE-E compared to Homogeneous Graph-based CSS. By analyzing these results, we find that the baseline models only model the current dialogue history, failing to leverage richer dialogue information in stored dialogue. In contrast, RADKA-CSS effectively integrates style knowledge from early dialogue fragments in stored dialogue that are similar to the current dialogue scene and style through multi-attribute retrieval. This enables a more comprehensive understanding of conversational style, enhancing the naturalness and stylistic consistency of the synthesized speech. Furthermore, RADKA-CSS employs a multi-granularity heterogeneous graph to augment the multi-source style knowledge, significantly improving its understanding of the current conversational style. This not only aligns the synthesized speech more closely with the current conversational style but also greatly enhances the quality of speech synthesis.

\subsection{Ablation Results}
To validate the contribution of each component in RADKA-CSS, we analyze the ablation results by removing different components, as shown in Table \ref{aexp}.

\textbf{Abl.1-Abl.4} validate the effectiveness of different style knowledge. Specifically, \textbf{Abl.1} removes the retrieval of dialogues from SD. The subjective metrics N-DMOS and S-DMOS decrease by 0.197 and 0.185, respectively. The objective metrics MAE-P, MAE-E, and MAE-D drop by 0.021, 0.018, and 0.009, respectively. This result indicates that RADKA-CSS extracts style knowledge aligned with the current conversational style from the early dialogue fragments retrieved from SD, helping the agent better understand the style of the CD. \textbf{Abl.2} removes semantic modeling of the text modality in the CD. The subjective metrics N-DMOS and S-DMOS decrease by 0.135 and 0.139, respectively. The objective metrics MAE-P, MAE-E, and MAE-D drop by 0.011, 0.010, and 0.002, respectively. These results suggest that the dialogue semantics of the CD help the agent generate speech aligned with the current conversational style, improving the naturalness and fluency of the synthesized speech. \textbf{Abl.3} removes style modeling of the audio modality in the CD. Both subjective metrics N-DMOS and S-DMOS decrease by 0.178. The objective metrics MAE-P, MAE-E, and MAE-D drop by 0.016, 0.012, and 0.002, respectively. These findings highlight the effectiveness of the dialogue style in the CD and its importance in enhancing RADKA-CSS’s understanding of the current conversational style. \textbf{Abl.4} removes the predicted $a_N$ style vector. The subjective metrics N-DMOS and S-DMOS decrease by 0.129 and 0.128, respectively. The objective metrics MAE-P, MAE-E, and MAE-D drop by 0.014, 0.009, and 0.001, respectively. These results suggest that the $a_N$ style vector contains style information specific to the speech to be synthesized. In summary, the experiments in Abl.1-4 validate the contributions of the four types of style knowledge to RADKA-CSS. Abl.1 shows the largest metric declines, further demonstrating that retrieving dialogues with styles similar to the current dialogue scenario significantly helps the agent better understand the current conversational style. Conversely, Abl.4 shows the smallest metric declines, indicating that the $a_N$ style vector primarily contains information about the predicted $a_N$ style, with limited information about the dialogue scenario style.

\textbf{Abl.5-Abl.8} validate the effectiveness of different technical components. \textbf{Abl.5} replaces the heterogeneous graph with a homogenous graph. The subjective metrics N-DMOS and S-DMOS decrease by 0.172 and 0.155, respectively. The objective metrics MAE-P, MAE-E, and MAE-D drop by 0.029, 0.022, and 0.010, respectively. This demonstrates that the proposed heterogeneous graph structure, which includes three granularity levels of nodes and their four relationships, effectively enhances the feature expression of dialogue style and semantics. This enables RADKA-CSS to better capture and model the style and semantics of the dialogue. \textbf{Abl.6} removes the word-level and dialogue-level node features, leaving only the sentence-level nodes. The subjective metrics N-DMOS and S-DMOS decrease by 0.219 and 0.196, respectively. The objective metrics MAE-P, MAE-E, and MAE-D drop by 0.031, 0.017, and 0.012, respectively. These results suggest that the multi-granularity node features provide a more comprehensive expression of style and semantics. By integrating word-level, sentence-level, and dialogue-level node features, RADKA-CSS can better understand and model the style and semantics of the dialogue. \textbf{Abl.7} replaces the knowledge aggregation method with direct addition. The subjective metrics N-DMOS and S-DMOS decrease by 0.150 and 0.156, respectively. The objective metrics MAE-P, MAE-E, and MAE-D drop by 0.023, 0.006, and 0.003, respectively. This indicates that directly adding different style knowledge representations results in redundancy in the style information, weakening the model’s ability to capture key style features and impacting its ability to render the dialogue style. \textbf{Abl.8} removes retrieval-based dialogue contrastive learning. The subjective metrics N-DMOS and S-DMOS decrease by 0.147 and 0.180, respectively. The objective metrics MAE-P, MAE-E, and MAE-D drop by 0.016, 0.013, and 0.001, respectively. This indicates that contrastive learning brings the semantic and style representations of similar dialogues closer. At the same time, it further distinguishes the semantic and style representations of dialogues with different scenarios and inconsistent styles, enhancing the model's ability to capture and differentiate both conversational style and semantics. Abl.5-8 validate the contributions of different technical components to RADKA-CSS. The multi-granularity heterogeneous graph modeling effectively encodes the semantics and style of the dialogue, while a proper knowledge aggregation method and the use of contrastive learning to constrain the retrieval of conversational style and semantic representations are crucial for the quality and style of the agent's synthesized speech.

It is important to note that, in \textbf{Abl.9}, we replace the retrieved Top-K dialogues with the ground truth Top-K dialogues, leading to most experimental metrics outperforming those of our proposed RADKA-CSS. Compared to our proposed RADKA-CSS, the subjective metrics N-DMOS (3.962) and S-DMOS (3.906) improve by 0.058 and 0.027, respectively. The objective metrics MAE-P (0.440) and MAE-E (0.301) improve by 0.002 and 0.004, respectively. This indicates that using dialogue style knowledge more similar to the CD to help the agent synthesize speech can significantly improve speech quality and style performance.

\begin{table}[t!]
\centering
\caption{The CMOS results of different retrieval schemes.}
\begin{tabular}{ccccccc}
\hline
  A vs B &  &  & \textbf{N-CMOS} ($\uparrow$)  &  & \textbf{S-CMOS} ($\uparrow$)  & \\ \hline

Rs.1 vs Rs.2 &  &  & 0.312   &  & 0.377  &   \\

Rs.1 vs Rs.3 &  &  & 0.250   &  & 0.314  &   \\

Rs.1 vs Rs.4 &  &  & 0.375   &  & 0.621  &   \\

Rs.1 vs Rs.5 &  &  & 0.442   &  & 0.685  &   \\

Rs.1 vs Rs.6 &  &  & 0.662   &  & 0.881  &   \\

Rs.1 vs Rs.7 &  &  & -0.208   &  & -0.215  &   \\  \hline

\end{tabular}
\label{cmosexp}
\end{table}
\vspace{-3mm}

% -----------------------

\begin{table}[t!]
\centering

\caption{The Recall results of different retrieval schemes.}
\begin{tabular}{ccccccc}
\hline
Rs.x & \textbf{R@1} & \textbf{R@2} & \textbf{R@3}  & \textbf{R@4} & \textbf{R@5} &  \textbf{R@10}  \\ \hline

\textbf{Rs.1} & \textbf{0.450} & \textbf{0.500} & \textbf{0.667} & \textbf{0.750} & \textbf{0.800} & \textbf{0.900} \\

Rs.2 & 0.437 & 0.487 & 0.645 & 0.726 & 0.773 & 0.872 \\

Rs.3 & 0.441 & 0.495 & 0.658 & 0.740 & 0.789 & 0.889 \\

Rs.4 & 0.425 & 0.481 & 0.634 & 0.712 & 0.758 & 0.857 \\

Rs.5 & 0.430 & 0.483 & 0.638 & 0.716 & 0.764 & 0.863\\ \hline

\end{tabular}

\label{reexp}
\end{table}

% -------------------------------------

\subsection{Analysis of Retrieval Schemes}
To validate the effectiveness of our proposed multi-attribute retrieval method, we conduct both subjective and objective experiments, comparing it with six other retrieval schemes. CMOS is used as the subjective evaluation metric, and Recall is used as the objective. These retrieval schemes are: \textbf{1) Retrieval Scheme 1 (Rs.1):} sums the dialogue semantic similarity and dialogue style similarity, then selects Top-K based on the combined similarity; \textbf{2) Retrieval Scheme 2 (Rs.2):} first calculates the dialogue semantic similarity, then calculates the dialogue style similarity and selects Top-K; \textbf{3) Retrieval Scheme 3 (Rs.3):} first calculates the dialogue style similarity, then calculates the dialogue semantic similarity and selects Top-K; \textbf{4) Retrieval Scheme 4 (Rs.4):} uses only dialogue semantic similarity to select Top-K; \textbf{5) Retrieval Scheme 5 (Rs.5):} uses only dialogue style similarity to select Top-K; \textbf{6) Retrieval Scheme 6 (Rs.6):} selects Top-K randomly; \textbf{7) Retrieval Scheme 7 (Rs.7):} uses the ground truth Top-K.

For subjective results, as shown in Table \ref{cmosexp}. The speech synthesized by RADKA-CSS using Rs.1 outperforms the speech synthesized using Rs.2-Rs.6 in both naturalness and style, though it is not as good as Rs.7. 

\textbf{Rs.1-Rs.3} belong to the multi-attribute retrieval scheme. Specifically, comparing Rs.1 with Rs.2 (\textbf{Rs.1 vs Rs.2}), the N-CMOS score is 0.312, and the S-CMOS score is 0.377. This indicates that the speech synthesized by Rs.1 is preferred over Rs.2 in both naturalness and style. This is because Rs.1 considers both dialogue semantics and style, and the retrieved Top-K most likely ensures both similar dialogue style and scenario. However, Rs.2 first retrieves based on dialogue semantics, which may result in selecting dialogues with similar scenarios but vastly different styles, making it unable to filter dialogues with a style similar to the current conversation. Next, we compare Rs.1 with Rs.3 (\textbf{Rs.1 vs Rs.3}), with N-CMOS and S-CMOS scores of 0.250 and 0.314, respectively. The speech synthesized by Rs.1 is again preferred over Rs.3 in both naturalness and style. Similarly, Rs.3 first retrieves based on dialogue style, which may exclude dialogues with similar styles but mismatched scenarios, failing to ensure overall similarity with the current dialogue. 

\textbf{Rs.4-Rs.5} belong to the single-attribute retrieval scheme. Comparing Rs.1 with Rs.4 and Rs.5 (\textbf{Rs.1 vs Rs.4} and \textbf{Rs.1 vs Rs.5}), the N-CMOS and S-CMOS scores are 0.375 and 0.621, and 0.442 and 0.685, respectively. This indicates that the speech synthesized by Rs.1 is preferred over both Rs.4 and Rs.5 in terms of naturalness and style. This is because Rs.4 and Rs.5 only use dialogue semantics or dialogue style as the retrieval scheme, meaning the selected Top-K are either only similar in scenario or only in style, failing to ensure overall similarity with the current conversation. 

\textbf{Rs.6-Rs.7} use random Top-K and ground truth Top-K, respectively. We compare Rs.1 with Rs.6 (\textbf{Rs.1 vs Rs.6}), where the N-CMOS and S-CMOS scores are 0.662 and 0.881, respectively. The speech synthesized by Rs.6 is far inferior to Rs.1 in both naturalness and style. This is because Rs.6 randomly selects Top-K, resulting in dialogues that are inconsistent with both the scenario and style of the current conversation, making it ineffective in helping the agent understand the current conversation. It is worth noting that we compare Rs.1 with Rs.7 (\textbf{Rs.1 vs Rs.7}), where the N-CMOS and S-CMOS scores are -0.208 and -0.215, respectively. This indicates that a better Top-K selection can guide RADKA-CSS to synthesize speech with higher naturalness and better style consistency.

Through the comparison of Rs.1 with Rs.2-Rs.6, we conclude that Rs.1 considers both dialogue semantics and style, so the selected Top-K is superior to the other schemes. The experiment comparing Rs.1 with Rs.7 demonstrates that selecting dialogues that are more similar to the current dialogue in both dialogue scenarios and styles leads to better RADKA-CSS performance.

For the objective evaluation, as shown in Table \ref{reexp}, Rs.1 achieves the highest recall rates when retrieving the top 1, top 5, and top 10 similar dialogues: R@1 (0.450), R@2 (0.500), R@3 (0.667), R@4 (0.750), R@5 (0.800), and R@10 (0.900), consistently outperforming Rs.2 through Rs.5. These results consistent with the subjective comparison findings, supporting our conjecture that the method used in Rs.1, which incorporates both dialogue semantics and style, obtains retrieval Top-K that are more similar to the ground truth Top-K. In contrast, Rs.2–Rs.5 fail to ensure comprehensive similarity with the current dialogue in terms of both semantics and style. Since Rs.6 randomly selects Top-K dialogues and Rs.7 uses the ground truth Top-K, we do not calculate recall for these two methods.

Rs.1 considers both dialogue semantics and style, making the selected Top-K superior to Rs.2–Rs.6. Therefore, we select Rs.1 as the retrieval scheme for RADKA-CSS.

%  -----------------------

\begin{figure}
    \centering
    \includegraphics[width=1\linewidth]{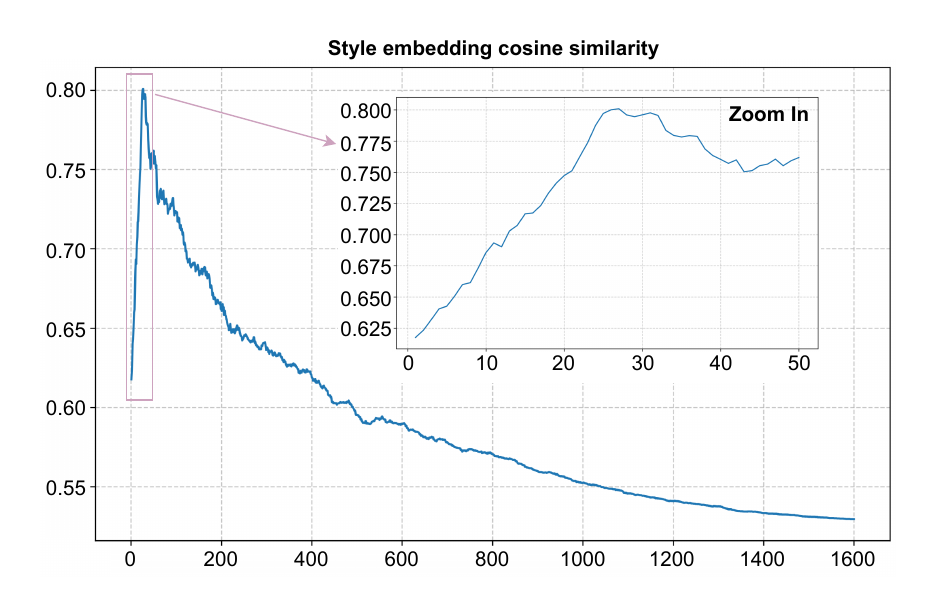}
    \vspace{-3mm}
    \caption{Style similarity between the fused retrieved Z-group dialogue style embeddings and the ground-truth style embedding.} \vspace{-3mm}
    \label{sim}
\end{figure}

% --------------------------

\subsection{Analysis about Selection of Z}
In Fig. \ref{sim}, we examine the similarity between the final style embedding ($FS_{emb}$), which is weighted by the Z sets of dialogue styles retrieved from SD during inference, and the ground-truth style embedding of the current dialogue. The results show that as the value of Z increases, the cosine similarity between the weighted $FS_{emb}$ and the ground-truth dialogue style embedding first increases and then decreases. Specifically, when Z is set to 1, the similarity between $FS_{emb}$ and the ground-truth style embedding is 0.617, and when Z is set to 10, the similarity reaches 0.685. When Z ranges from 1 to 25, the similarity exhibits an overall upward trend, reaching 0.797 when Z is set to 25. When Z is between 25 and 32, the similarity peaks. However, when the value of Z exceeds 32, the similarity gradually decreases. Considering the increased computational complexity with higher Z values, we select a Z value of 25 as the retrieval quantity during inference to balance performance and complexity.

\begin{figure}
    \centering
    \includegraphics[width=1\linewidth]{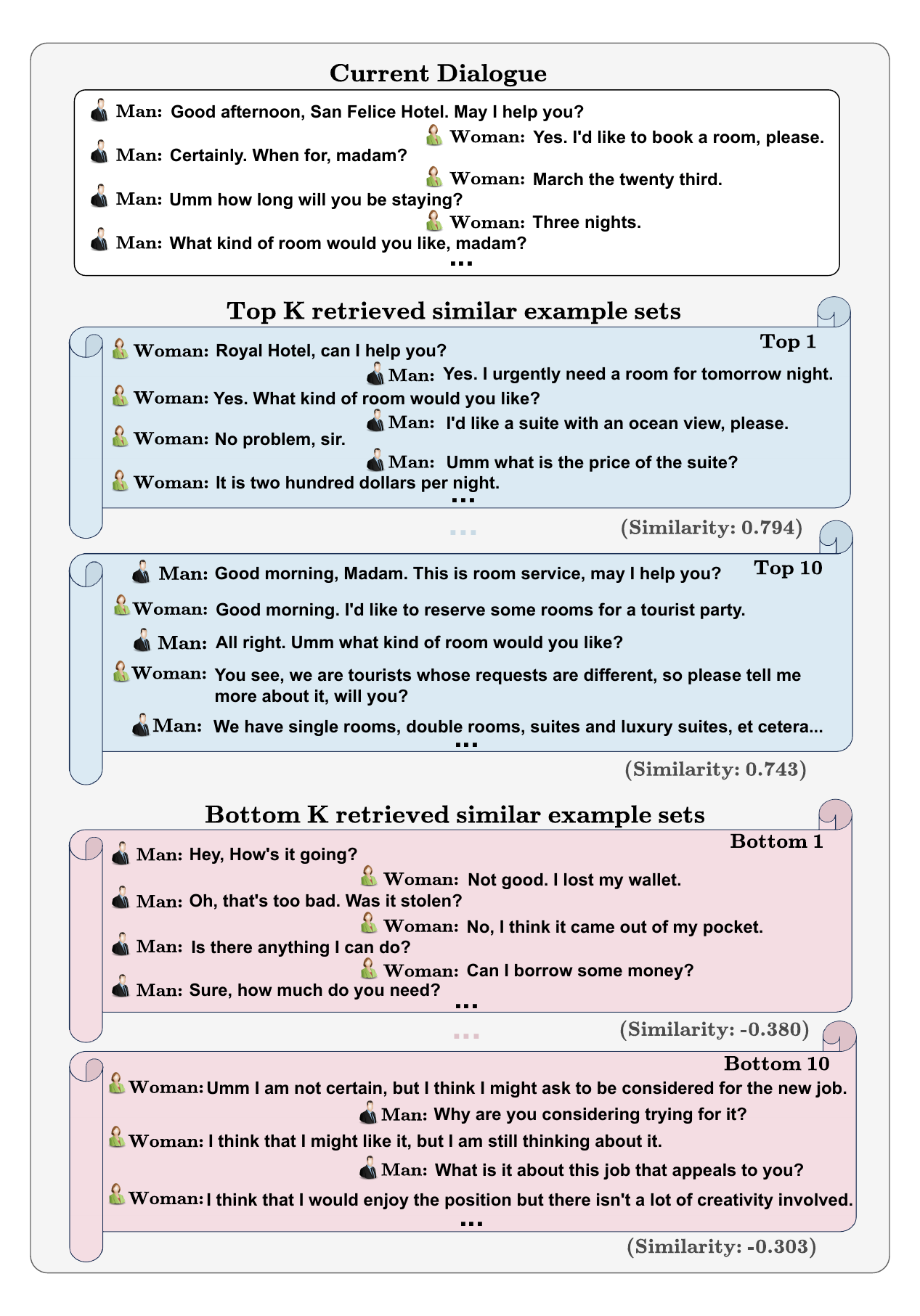}
    \vspace{-3mm}
    \caption{Examples of retrieved samples. The complete dialogue text and audio can be found at: \url{https://coder-jzq.github.io/RADKA-CSS-Website/index.html\#case-study}.} \vspace{-3mm}
    \label{casestudy}
\end{figure}

% -----------------------

\subsection{Case Study on Retrieval Content}
To validate that the RADKA-CSS retrieves dialogues from SD that are similar to CD in terms of both dialogue scenarios and style, we present CD along with the top 1, top 2, bottom 1, and bottom 10 dialogues ranked by similarity, where the similarity is calculated as the mean of dialogue semantic similarity and dialogue style similarity. As shown in Fig. \ref{casestudy}, we display a portion of the key dialogue text. CD belongs to a dialogue scenario about hotel room reservations. The dialogues ranked in the Top 1 (similarity: 0.794) and Top 10 (similarity: 0.743) are all related to hotel room reservations, while those ranked in the Bottom 1 (similarity: -0.380) and Bottom 10 (similarity: -0.303) do not belong to the hotel reservation scenario. Based on the audio analysis of these dialogues, the dialogue styles in the Top 1 and Top 10 are very similar to CD, whereas the styles in the Bottom 1 and Bottom 10 show significant differences. The complete dialogue text and audio can be found at: \url{https://coder-jzq.github.io/RADKA-CSS-Website/index.html#case-study}.

\section{Conclusion}
\label{sec7}
To improve the ability of CSS systems to synthesize speech that aligns with the current conversational style, this paper proposes a novel RADKA-CSS model. We first build a database, SDSSD, that supports dialogue semantic and dialogue style retrieval. RADKA-CSS retrieves the Top-K dialogues from SDSSD that are similar to a CD in dialogue scenario and style through Multi-attribute Retrieval. Then, the Multi-source Style Knowledge Aggregator integrates style knowledge from CD, style knowledge from SD, and the predicted $a_N$ style knowledge to augment the style information. Finally, the Style Renderer effectively adds the augmented style knowledge into the linguistic encoding of the speech to be synthesized, helping to generate speech that aligns with the current conversational style and is expressive. Experimental results demonstrate the superiority of RADKA-CSS over state-of-the-art CSS systems. To the best of our knowledge, RADKA-CSS is the first study to apply retrieval-augmented generation to conversational speech synthesis. We hope our work will inspire further research in intelligent user-agent interaction. Additionally, although RADKA-CSS performs well in scenarios involving fixed-user and agent interactions, it does not currently support public-facing scenarios involving alternating interactions with multiple speakers.  In the future, we aim to address the challenges of extending RADKA-CSS to adapt to public-facing scenarios.

% \appendix
% \section{My Appendix}
% Appendix sections are coded under \verb+\appendix+.

% \verb+\printcredits+ command is used after appendix sections to list 
% author credit taxonomy contribution roles tagged using \verb+\credit+ 
% in frontmatter.
% \verb+\printcredits+

% \noindent \textbf{\Large Declaration of competing interest}
% xxx.

% \noindent \textbf{\Large Data availability}
% xxx.

% \noindent \textbf{\Large Acknowledgments }
% xxx.

% %% Loading bibliography style file
% %\bibliographystyle{model1-num-names}
% \bibliographystyle{cas-model2-names}

\bibliography{cas-refs}

\end{sloppypar}
\end{document}